\documentclass[runningheads]{llncs}

\usepackage{eccv}

\usepackage{eccvabbrv}

\usepackage{graphicx}
\usepackage{booktabs}
\usepackage{amsmath,amssymb}
\usepackage{bm}
\usepackage{booktabs}
\usepackage{multirow}
\usepackage{wrapfig}
\usepackage {colortbl,array,xcolor}

\usepackage[accsupp]{axessibility}  

\usepackage{hyperref}

\usepackage{orcidlink}

\begin{document}

\title{Accuracy Improvement of Cell Image Segmentation Using Feedback Former} 

\titlerunning{Feedback Former}

\author{Hinako Mitsuoka\inst{1}\orcidlink{0009-0005-6969-4017} \and
Kazuhiro Hotta\inst{1}\orcidlink{0000-0002-5675-8713}}

\authorrunning{H. Mitsuoka et al.}

\institute{Meijo University, 1-501 Shiogamaguchi, Tempaku-ku, Nagoya 468-8502, Japan\\
\email{200442165@ccalumni.meijo-u.ac.jp}, 
\email{kazuhotta@meijo-u.ac.jp}}

\maketitle

\begin{abstract}
    Semantic segmentation of microscopy cell images by deep learning is a significant technique. We considered that the Transformers, which have recently outperformed CNNs in image recognition, could also be improved and developed for cell image segmentation.
    Transformers tend to focus more on contextual information than on detailed information.
    This tendency leads to a lack of detailed information for segmentation.
    Therefore, to supplement or reinforce the missing detailed information, we hypothesized that feedback processing in the human visual cortex should be effective.
    Our proposed Feedback Former is a novel architecture for semantic segmentation, in which Transformers is used as an encoder and has a feedback processing mechanism.
    Feature maps with detailed information are fed back to the lower layers from near the output of the model to compensate for the lack of detailed information which is the weakness of Transformers and improve the segmentation accuracy.
    By experiments on three cell image datasets, we confirmed that our method surpasses methods without feedback, demonstrating its superior accuracy in cell image segmentation.
    Our method achieved higher segmentation accuracy while consuming less computational cost than conventional feedback approaches.
    Moreover, our method offered superior precision without simply increasing the model size of Transformer encoder, demonstrating higher accuracy with lower computational cost.
    \textcolor{magenta}{
    This work has been published in IEEE Access with DOI: \url{https://doi.org/10.1109/ACCESS.2025.3552847}. 
    Please refer to the published version for the final authoritative version of the paper.
    }
\end{abstract}

\section{Introduction}
\label{sec:intro}

Semantic segmentation, the process of assigning a class label to each pixel in an image, is an essential technique in various applications including medicine \cite{unet, vnet}, cell biology \cite{clstms, livecell}, and scene understanding \cite{segnet, pspnet}.
To achieve better performance, it is important to appropriately capture both detailed features, which refer to small patterns within the image, and contextual features, which refer to the rough positions of objects.
These are believed to exist at different resolutions or scales in network models, and various methods have been proposed to give models the ability to learn contextual features between large pixels without missing important detailed features \cite{hrnet, deeplab, pspnet}.

Although CNNs are a fundamental method for image recognition, there have been remarkable developments in Transformers, including Vision Transformer (ViT) \cite{vit}, which was proposed as a different approach.
Recently, Transformers have been successful in outperforming CNNs in various tasks such as image classification \cite{swin}, object detection \cite{co-detr}, and semantic segmentation \cite{segformer, mask2former}.
Transformers excel at capturing contextual information rather than detailed information\cite{cnntexture, cnnandtf} and perform the best when pretrained on large datasets. However, Transformer's performance may be limited on cell images that contain microstructures and less data.
Therefore, CNN-based methods\cite{xnet, unet} with superior ability to obtain local information are mainly used in cell image segmentation.
We believe that Transformers, which have been successful in many recognition tasks, can be further improved and developed in cell image segmentation.
Thus, we analyze the weaknesses of Transformers in this task.
Since Transformers extracts features from the entire image in parallel, it has a wider receptive field and is better at representing contextual information and relationships.
However, the above characteristics are strengths of Transformers, and the fact that contextual information is prioritized and enough detailed information is not available can also be a disadvantage in segmentation tasks involving cell images.
Consequently, we consider the lack of detailed information to be a weakness of Transformers in this task and propose a method as one way to improve it.

We focus on feedback processing in the visual cortex of the human brain.
In the human brain, not only feedforward processing from the lower layer which handles low-level information to the upper layer which handles high-level information but also feedback processing from the upper layer to the lower layer is performed \cite{feedback1,feedback2,feedback3}.
Neural networks such as CNNs and Transformers are mathematical models of the neural circuit structure of the human brain \cite{suuriteki-model}.
Nonetheless, typical neural networks only perform feedforward processing from the lower layers to the upper layers.

This paper proposes \textbf{Feedback Former} that compensates for weaknesses and improves accuracy by combining Transformers with feedback processing inspired by human visual cortex.
We consider that the detailed and important information missing in Transformers is passed from the feature maps near the output to the feature maps at lower layers, allowing the model to make better decisions.
Our approach is the first method to introduce feedback processing into the Transformer-based segmentation model.
We introduce feedback processing to a segmentation model that uses Attention as the TokenMixer part in MetaFormer \cite{poolformer, metaformer} framework as Encoder and Semantic FPN \cite{semfpn} as Decoder.
When Attention is used for the TokenMixer part of MetaFormer framework, the model structure is similar to ViT and Pyramid Vision Transformer \cite{pvt}.
We verify the effectiveness of the combination of Transformers and feedback processing.

We evaluated our proposed method on three different cell image datasets.
The results showed that the proposed method improved the segmentation accuracy on three cell image datasets compared to the conventional methods without feedback processing and conventional feedback approaches \cite{fbattn}.
Furthermore, we evaluated our proposed method and showed that it is lighter and more accurate than conventional feedback processing modules and that our proposed method outperforms larger models of similar size in MACs.
Our main contributions are:
\begin{itemize}
    \setlength{\parskip}{0cm} 
    \setlength{\itemsep}{0cm} 
    \item We propose a novel model structure, Feedback Former, which combines Transformers and feedback processing inspired by the human visual cortex.
    \item To improve the performance of Transformers, the feedback processing supplements the detailed information required for segmentation.
    \item 
    Feedback Former improved the segmentation accuracy on three cell image datasets compared to the method without feedback processing, especially on the iRPE dataset \cite{irpe} by \textbf{4.54\%}.
\end{itemize}


\section{Related Works}
\label{sec:relatedwork}

\subsection{Semantic Segmentation}
\label{sub:semseg}

Segmentation models must properly classify both detailed and contextual features.
Many approaches use an encoder-decoder structure to obtain multi-scale information \cite{unet, segnet, vnet}. 
CNN-based and Transformer-based methods have been developed to obtain multi-scale information effectively.
While Transformer-based models \cite{segformer, setr, mask2former} achieve high accuracy in benchmarks, they are less used for cell image segmentation. 
This is because Transformer models struggle to capture the detailed information for classifying fine structures like cells.
Thus, we propose to supplement this detailed information through feedback processing, similar to that in the human brain \cite{feedback1,feedback2,feedback3}.

\subsection{Conventional methods using feedback}
\label{sub:feedback methods}

Neural networks imitate the human brain, particularly the workings of neurons and the visual cortex. 
The visual cortex has a layered structure, utilizing both feedforward and feedback processing \cite{feedback1,feedback2,feedback3}. 
However, traditional neural network models primarily only use feedforward processing. 
Incorporating feedback processing, a fundamental mechanism in the human visual cortex could significantly advance image recognition.

Recently, several methods using feedback processing have been proposed \cite{runet, fmunet, fbattn, lfbnet, afnet}.
Especially for cell and medical images, some methods have been proposed for segmentation tasks using CNN-based models such as U-Net \cite{unet}.
FM-Unet \cite{fmunet} introduced a feedback path in U-Net structure to solve the problems of information loss in the encoder and lack of information in the decoder.
However, because it consists only of a convolutional layer, the receptive field is localized, making it difficult to learn contextual information.
Feedback Attention \cite{fbattn} uses a Self Attention mechanism to highlight important information in feature maps near the model's output and combines them with lower-layer feature maps. 
The input image passes through the model twice, with only the second output used for loss computation and training. These feedback approaches can improve accuracy by allowing the model to focus on specific areas based on previous results.
\begin{figure*}[t]
    \centering
    \includegraphics[scale=0.33]{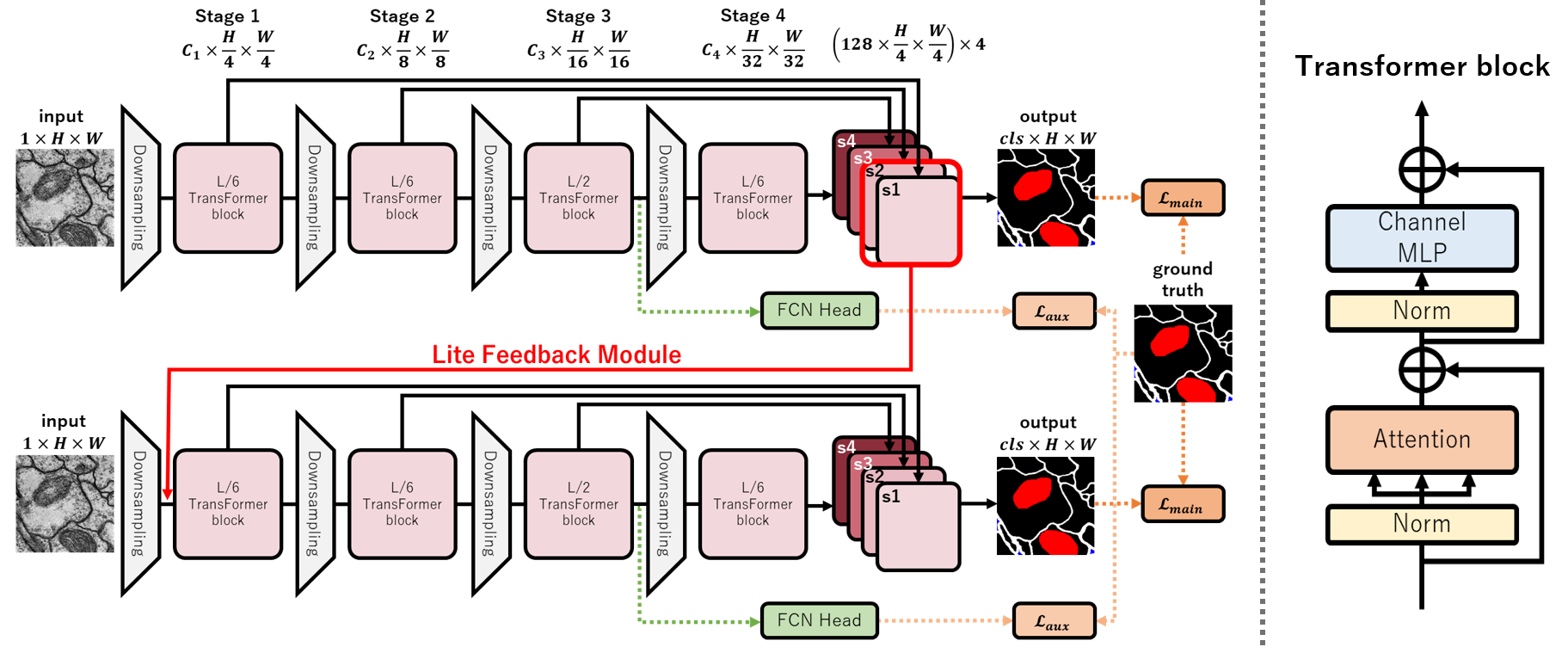}
    \caption{The overview of the architecture of the Feedback Former.
    }
    \label{fig:fbformer}
\end{figure*}
Methods that combine Transformers and feedback processing exist for other tasks, but not yet in segmentation task.
Thus, we introduce feedback processing influenced by the human brain and the above methods to the segmentation model using Transformers in the encoder to improve the segmentation accuracy.
\begin{wrapfigure}[11]{r}[0pt]{0.4\textwidth}
  \centering
  \includegraphics[scale=0.25]{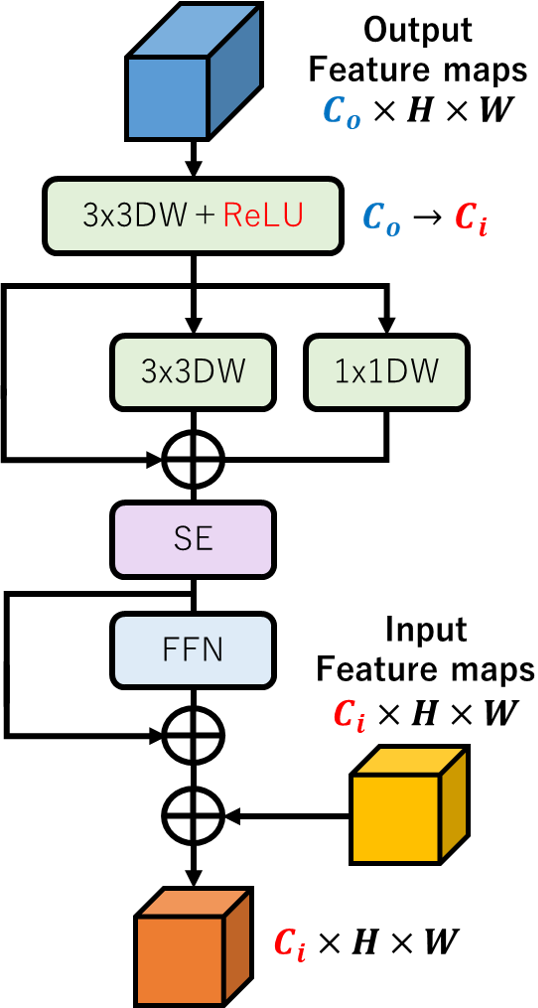}
  \caption{Lite Feedback Module}
  \label{fig:litefbmodule}
\end{wrapfigure}

\section{Proposed Method}
\label{sec:method}

\subsection{Feedback Former}
\label{sub:feformer}

\cref{fig:fbformer} presents the architecture of our proposed Feedback Former. 
Following \cite{poolformer}, we use MetaFormer as the encoder and Semantic FPN as the decoder for semantic segmentation. 
In this method, Attention is used in the TokenMixer part of the Metaformer framework, making the encoder of Feedback Former a Transformer.
The input image is downsampled to $\frac{H}{4} \times \frac{W}{4}$ at Stage 1 of MetaFormer and halved at each subsequent stage. 
Outputs from Stages 1, 2, 3, and 4 are input to Semantic FPN, convolved to produce feature maps s1, s2, s3, and s4 $\in \mathbb{R}^{128 \times \frac{H}{4} \times \frac{W}{4}}$.

In conventional methods, feature maps are combined and upsampled to match the input resolution as final output. 
In contrast, our method performs segmentation twice to incorporate feedback processing using feature maps near the model output. 
In the first round, feature maps s1 and s2 are concatenated and processed by the Lite Feedback Module, then passed to the lower layer of the second round.
The Lite Feedback Module, detailed in \cref{sub:litefeedback}, enhances important information in the feature maps fed back to the second round.
Since s1 and s2 are outputs from Stages 1 and 2 of MetaFormer, they contain relatively detailed features. 
By feeding them back, we address the Transformer's lack of detailed features. 
The output of the second round is the final output of our proposed method.

We describe the loss functions.
We employ the CE loss and the IoU loss \cite{iouloss}.
Following \cite{pspnet,pidnet}, we place an FCN head at the output of Stage 3 of MetaFormer to generate the auxiliary loss $\mathcal{L}_{aux}$ for better optimization.
The placement of an FCN head at the output of Stage 3 is based on that PSPNet \cite{pspnet} generates the auxiliary loss after res4b22 of ResNet101 \cite{resnet}.
FCN head is a simple decoder with two convolutional layers.
The loss of the entire model output is $\mathcal{L}_{main}$, and the loss for one round can be written as 
\begin{eqnarray}
    \label{eq:loss-koko}
    \mathcal{L}_{main} &=& \mathcal{L}_{aux} \;=\; \lambda_1 CELoss + \lambda_2 IoULoss  \\
    \label{eq:loss-par-lap}
    \mathcal{L} &=& \mathcal{L}_{main} + \lambda_3 \mathcal{L}_{aux} 
\end{eqnarray}
The parameters were set to $\lambda_1 = 0.7$, $\lambda_2 = 0.3$, and $\lambda_3 = 0.4$.
In addition, the proposed method uses both the first and second-round losses to update parameters.
Therefore, the final loss can be expressed by the first round loss $\mathcal{L}_{1st}$, the second round loss $\mathcal{L}_{2nd}$, and Equation (\ref{eq:loss-par-lap}).
\begin{eqnarray}
    \mathcal{L}_{1st} &=& \mathcal{L}_{2nd} \;=\; \mathcal{L} \label{eq:loss-12lap} \\
    Loss &=& \alpha \mathcal{L}_{1st} + \mathcal{L}_{2nd} \label{eq:loss-all}
\end{eqnarray}
where $\alpha$ is a hyperparameter.
In this paper, we set $\alpha$ to 0.5.
This is because the quality of information fed back to the second round is improved by ensuring segmentation accuracy in the first round.

\subsection{Lite Feedback Module}
\label{sub:litefeedback}

We propose Lite Feedback Module as illustrated in \cref{fig:litefbmodule}. 
This lighter module aims to improve the accuracy of Transformer-based encoders. 
The feature maps s1 and s2 are concatenated at the first round of Feedback Former, and we process them to emphasize important information, and add them to the lower layer at the second round.
Transformers lack detailed information, so we extract and emphasize details of the feature map in the first round and feed it back to the second round. 
Attention mechanisms used in conventional methods with feedback processing can extract important information but are computationally expensive.
Instead, a convolution that processes local neighborhoods may work well for feedback processing, as it efficiently supplements detailed information.

Therefore, we use Depthwise Convolution (DW Conv) to extract detailed information while reducing computational cost. 
Lite Feedback Module, inspired by \cite{repvit}, combines light computation with high accuracy. 
It efficiently mixes information spatially using DW Conv in the first half and mixes it in the channel direction in the second half.
The enhanced feature map from the first round is then multiplied by a learnable parameter $\beta$ (initial value 1) and added to the second round's lower layer feature map.

In summary, Lite Feedback Module extracts important information from the first round and passes it to the second round, enabling the model to make more accurate decisions.

\section{Experiments}
\label{sec:experiments}

\subsection{Datasets and Metrics}
\label{sub:dataset}

In experiments, we used three datasets. 
The first dataset is Drosophila cell image dataset (Drosophila) \cite{drosophila} which consists of 5 classes: membrane, mitochondria, synapse, glia/extracellular, and intracellular.
The original images are $1024 \times 1024$, but we cropped them to $256 \times 256$ due to GPU memory limitations. 
There are 320 non-overlapping cropping areas in total, with 192 regions for training, 64 for validation, and 64 for inference. 
We evaluated our method using 5-fold cross-validation.
Although the same dataset is used in \cite{fbattn}, they used only 4 classes in all 5 classes. Thus, the paper \cite{fbattn} gave a higher accuracy than ours.
This dataset is licensed under CC BY-NC-SA 3.0.

The second dataset is 2D electron microscopy images in ISBI2012 challenge (ISBI2012) \cite{isbi}, consisting of 2 classes: membrane and background. 
The original images are $512 \times 512$, but we cropped them to $256 \times 256$ due to GPU memory limitations. 
There are 120 non-overlapping cropping areas in total.
The third dataset is absorbance microscopy images of human iRPE cells (iRPE) \cite{irpe}, consisting of 2 classes: membrane and background. 
The dataset contains 1,032 images. 

For the last two datasets, we randomly rearranged images and split each dataset into training and inference data in a 2:1 ratio. 
We used 3-fold cross-validation by changing training and inference data.
Intersection over Union (IoU) per class and mean IoU (mIoU) were used as evaluation metrics.

\subsection{Training conditions}
\label{sub:implementation}

We used Pytorch library and trained the model with Adam for $500$ epochs, using a batch size of 4 on an Nvidia GTX1080Ti. The learning rate started at $0.001$ and gradually decreased using a cosine scheduler \cite{cosinelr}. 
The model with the highest mIoU during validation was saved for testing. The loss function described in \cref{sub:feformer} was used. 
For data pre-processing, training data were randomly flipped vertically or horizontally and rotated by $\pm 90$ degrees.

In the following experiments, we compared segmentation models using the MetaFormer framework's encoder. 
We used Semantic FPN \cite{semfpn} as the common decoder and AttentionFormer as encoders. MetaFormer with Attention Tokenmixer is referred to as AttentionFormer. 
We introduced the Feedback Former architecture to models with the AttentionFormer encoder, resulting in four comparison methods: our proposed Feedback Former, two Feedback Attention approaches \cite{fbattn} (using Self Attention and Source-Target Attention), and a method without feedback processing. 
For Feedback Attention, we used the same model and feature maps as the proposed method to compare the feedback processing module. 
The encoder size was S12 as indicated in \cite{metaformer, poolformer}.

\begin{table*}[t]
    \centering
    \caption{Accuracy on the Drosophila dataset. In Feedback Attention, ST indicates Source-Target Attention, Self indicates Self Attention.}
    \scalebox{0.74}{
    \begin{tabular}{lcccccc}
    \toprule
    Methods & membrane & mitochondria & synapse & 
    \begin{tabular}{c}
    glia/\\extracellular 
    \end{tabular}
    & intracellular & mIoU\\     
    \hline\hline
        AttentionFormer-S12
        &69.25&77.45&49.61&62.81&91.67&70.16 \\
        Feedback\\Attention(ST)\cite{fbattn}
        &68.11&75.84&48.79&61.84&91.37&69.19 \\
        Feedback\\Attention(Self)\cite{fbattn}
        &70.47&78.81&48.48&65.94&91.89&71.12 \\
        \rowcolor[gray]{0.85}
        Feedback Former
        &\textbf{71.78}&\textbf{80.50}&\textbf{51.31}&\textbf{67.19}&\textbf{92.31}&\textbf{72.62} \\
    \bottomrule
    \label{tab:drosophila}
    \end{tabular}    
    }
\begin{minipage}[t]{0.49\hsize}
    \centering
    \caption{Accuracy on ISBI2012 dataset.}
    \scalebox{0.74}{
    \begin{tabular}{lccc}    
    \toprule
    Methods & membrane & background & mIoU\\     
    \hline\hline
        AttentionFormer-S12
        &52.65&83.22&67.93 \\
        Feedback\\Attention(ST)\cite{fbattn}
        &50.44&82.43&66.43 \\
        Feedback\\Attention(Self)\cite{fbattn}
        &54.23&83.90&69.06 \\
        \rowcolor[gray]{0.85}
        Feedback Former 
        &\textbf{56.45}&\textbf{85.22}&\textbf{70.84} \\
    \bottomrule 
    \label{tab:isbi2012}
    \end{tabular} 
    }
\end{minipage}
\hfill
\begin{minipage}[t]{0.49\hsize}
    \centering
    \caption{Accuracy on iRPE dataset.}
    \scalebox{0.74}{
    \begin{tabular}{lccc}
    \toprule
    Methods & membrane & background & mIoU\\     
    \hline\hline
        AttentionFormer-S12
        &41.74&72.05&56.89 \\
        Feedback\\Attention(ST)\cite{fbattn}
        &44.30&72.81&58.55 \\
        Feedback\\Attention(Self)\cite{fbattn}
        &45.90&73.48&59.69 \\
        \rowcolor[gray]{0.85}
        Feedback Former 
        &\textbf{48.71}&\textbf{74.16}&\textbf{61.43} \\
    \bottomrule
    \end{tabular}
    \label{tab:irpe}
    }
\end{minipage}
\end{table*}

\subsection{Experimental results}
\label{sub:experimantal-results}

\subsubsection{Quantitative results}
\label{subsub:teiryou}

\cref{tab:drosophila}, \ref{tab:isbi2012}, and \ref{tab:irpe} show comparison results on the three datasets.
The proposed method is more accurate than conventional methods and those without feedback processing. 
By using the Lite Feedback Module, detailed features from the first round's output are emphasized and passed to the second round, improving accuracy. 
Unlike conventional methods, our approach shows that an Attention mechanism is not necessarily required to emphasize important information.
The proposed method improved the segmentation accuracy of AttentionFormer on all three datasets: 2.46\% for Drosophila, 2.91\% for ISBI2012, and 4.54\% for iRPE. 
These results indicate that feedback processing is effective for Transformer-based models like AttentionFormer and demonstrate that Feedback Former is versatile and effective across multiple cell image datasets.

\subsubsection{Qualitative results}
\label{subsub:teisei}

\begin{figure*}[t]
    \centering
    \includegraphics[scale=0.3]{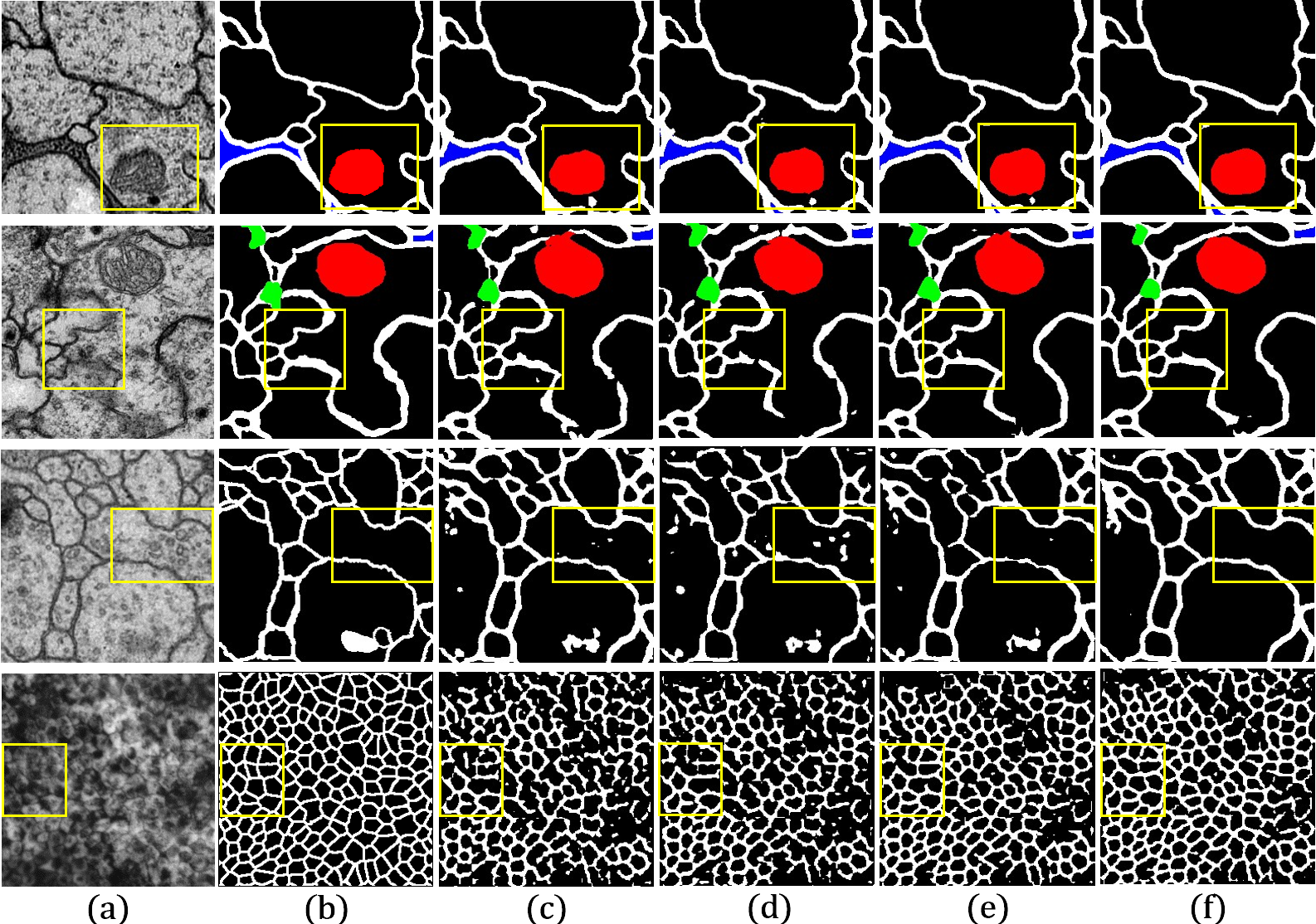}
    \caption{Qualitative results. The first and second rows are the results on Drosophila, the third row is the results on ISBI2012, and the bottom row is the result on iRPE dataset. (a) Input image, (b) Ground truth, (c) AttentionFormer, (d) Feedback Attention(ST) \cite{fbattn}, (e) Feedback Attention(Self) \cite{fbattn}, (f) Feedback Former. 
    }
    \label{fig:result-qualitative}
\end{figure*}

\cref{fig:result-qualitative} shows the segmentation results by each method on the three datasets.
The yellow frames in \cref{fig:result-qualitative} show that the proposed method improved the segmentation of membrane (white), which were excessively segmented by the conventional methods on the Drosophila dataset.
Furthermore, on the ISBI2012 and iRPE datasets, the proposed method improved the membrane (white) fragmentation and over-detection observed with the conventional methods.
This indicates that the proposed method's ability to classify cell membranes, which require detailed information, especially in cell images, has been improved by the appropriate feedback of detailed information.

\subsubsection{Verification of computational complexity and efficiency}
\label{subsub:keisan}

As shown in \cref{tab:macs}, we compared the MACs of our Lite Feedback Module with the feedback module of conventional Feedback Attention \cite{fbattn}. 
Lite Feedback Module reduces MACs to about 36\% of conventional method. 
This advantage comes from using lightweight CNN structures like DW Conv instead of the computationally expensive Attention mechanism.
Segmentation accuracy using Lite Feedback Module is also higher than with the conventional feedback module. 
This indicates that for Transformer-based encoders, it is sufficient to enhance detailed information using a CNN-based module. 
Thus, our Lite Feedback Module is a lightweight and high-performance module suitable for Transformer-based encoders.

As shown in \cref{tab:large}, we compared MACs and the accuracy of Transformer-based encoder by simply increasing the model size.
These results indicate that Feedback Former is more accurate for AttentionFormer-S36, which has the same level of MACs as the overall Feedback Former.
The result shows that the proposed method can improve accuracy more efficiently than simply increasing the size of the Transformer-based encoder.

\begin{table*}[t]
\begin{minipage}[t]{0.49\hsize}
    \centering
    \caption{Comparison of MACs of the feedback module on the Drosophila dataset.}
    \scalebox{0.74}{
    \begin{tabular}{lccc}    
    \toprule
    Modules & Params(M) & MACs(G) & mIoU\\     
    \hline\hline
        Feedback\\Attention(ST)\cite{fbattn}
        &0.05&0.21&69.19 \\
        Feedback\\Attention(Self)\cite{fbattn}
        &0.06&0.23&71.12 \\
        \rowcolor[gray]{0.85}
        Lite Feedback Module 
        &\textbf{0.02}&\textbf{0.08}&\textbf{72.62} \\
    \bottomrule 
    \label{tab:macs}
    \end{tabular} 
    }
\end{minipage}
\hfill
\begin{minipage}[t]{0.49\hsize}
    \centering
    \caption{Comparison of MACs with larger size encoders on the Drosophila dataset.}
    \scalebox{0.74}{
    \begin{tabular}{lccc}
    \toprule
    Encoders & Params(M) & MACs(G) & mIoU\\     
    \hline\hline
        AttentionFormer-S12
        &18.13&5.74&70.16 \\
        AttentionFormer-S24
        &32.32&8.87&69.82 \\
        AttentionFormer-S36
        &46.48&11.93&70.69 \\
        \rowcolor[gray]{0.85}
        Feedback Former
        & & & \\
        \rowcolor[gray]{0.85}
        (AttentionFormer-S12)
        &36.28&11.56&\textbf{72.62} \\
    \bottomrule
    \end{tabular}
    \label{tab:large}
    }
\end{minipage}
\end{table*}

\section{Conclusion}
\label{sec:conclusion}

We proposed Feedback Former to enhance significant features by feeding back the network's output to lower layers, similar to the human brain. 
We also introduced the Lite Feedback Module, which is more efficient than conventional feedback methods, and demonstrated its effectiveness with quantitative results. 
The proposed method improved the segmentation accuracy of Transformer-based model more efficiently than simply increasing model size.



%
%
\bibliographystyle{splncs04}
\bibliography{main}
\end{document}